\documentclass[final]{cvpr}

\usepackage{times}
\usepackage{graphicx}
\usepackage{amssymb}
\usepackage{epsfig}
\usepackage{amsmath}

\usepackage{verbatim}
\usepackage{helvet}
\usepackage{courier}
\usepackage{amsfonts}
\usepackage{subfigure}
\usepackage{caption}
\graphicspath{ {./images/} }
\usepackage{booktabs}
\usepackage{threeparttable}
\usepackage{multicol}
\usepackage{multirow}
\usepackage{tabularx}
\newcolumntype{L}[1]{>{\raggedright\arraybackslash}p{#1}}
\newcolumntype{C}[1]{>{\centering\arraybackslash}p{#1}}
\newcolumntype{R}[1]{>{\raggedleft\arraybackslash}p{#1}}

\DeclareMathAlphabet\mathbfcal{OMS}{cmsy}{}{n}

\usepackage[pagebackref=true,breaklinks=true,letterpaper=true,colorlinks,bookmarks=false]{hyperref}

\def\cvprPaperID{5934} 
\def\confYear{CVPR 2021}

\begin{document}

\title{FFB6D: A Full Flow Bidirectional Fusion Network for 6D Pose Estimation}

\author{Yisheng He$^{1}$ \quad {Haibin Huang}$^{3}$ \quad {Haoqiang Fan}$^{2}$ \quad {Qifeng Chen}$^{1}$ \quad {Jian Sun}$^{2}$ \\
	${^1}$Hong Kong University of Science and Technology \quad \quad
	${^2}$ Megvii Technology \quad \quad 
	${^3}$Kuaishou Technology \\
}
\maketitle

\begin{abstract}

In this work, we present FFB6D, a \textbf{F}ull \textbf{F}low \textbf{B}idirectional fusion network designed for \textbf{6D} pose estimation from a single RGBD image. Our key insight is that appearance information in the RGB image and geometry information from the depth image are two complementary data sources, and it still remains unknown how to fully leverage them. Towards this end, we propose FFB6D, which learns to combine appearance and geometry information for representation learning as well as output representation selection. Specifically, at the representation learning stage, we build \textbf{bidirectional} fusion modules in the \textbf{full flow} of the two networks, where fusion is applied to each encoding and decoding layer. In this way, the two networks can leverage local and global complementary information from the other one to obtain better representations. Moreover, at the output representation stage, we designed a simple but effective 3D keypoints selection algorithm considering the texture and geometry information of objects, which simplifies keypoint localization for precise pose estimation. Experimental results show that our method outperforms the state-of-the-art by large margins on several benchmarks. Code and video are available at \url{https://github.com/ethnhe/FFB6D.git}.



\end{abstract}
\section{Introduction}

6D Pose Estimation is an important component in lots of real-world applications, such as augmented reality \cite{marchand2015pose}, autonomous driving \cite{geiger2012we,chen2017multi,xu2018pointfusion} and robotic grasping \cite{collet2011moped,tremblay2018deep,he2020pvn3d}. It has been proven a challenging problem due to sensor noise, varying lighting, and occlusion of scenes. Recently, the dramatic growth of deep learning techniques motivates several works to tackle this problem using convolution neural networks (CNNs) on RGB images \cite{xiang2017posecnn,peng2019pvnet,Zakharov2019dpod,li2019cdpn}. However, the loss of geometry information caused by perspective projection limits the performance of these approaches in challenging scenarios, such as poor lighting conditions, low-contrast scenes, and textureless objects. The recent advent of inexpensive  RGBD sensors provides extra depth information to ease the problem \cite{calli2015ycb,hinterstoisser2011multimodal,hodan2018bop,hodan2017tless} and also leads to an interesting research question: How to fully leverage the two data modalities effectively for better 6D pose estimation?

\newcommand{\cmpMDL}{1}
\begin{figure}
    \centering
    \subfigure[\textbf{The DenseFusion \cite{wang2019densefusion} Network}. The two networks extract features from different modalities of data separately without any communication, util the final layers of the encoding-decoding architecture.]{
        \begin{minipage}[]{\cmpMDL\linewidth}
            \centering
            \label{fig:dense_fusion}
            \includegraphics[scale=.75]{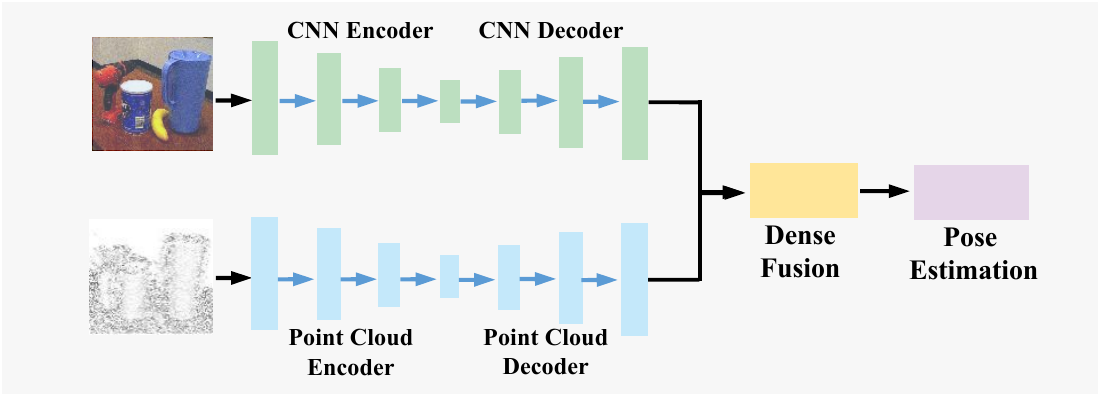}\\
        \end{minipage}%
    }%
    
    \subfigure[\textbf{The Proposed Full Flow Bidirectional Fusion Network}. Bidirectional fusion modules are added as bridges for information communication in the full flow of the two networks, where fusion is applied on each encoding and decoding layers. Local and global supplementary information from each other is shared between the two networks for better appearance and geometry representation learning, which is crucial for 6D pose estimation.]{
        \begin{minipage}[]{\cmpMDL\linewidth}
            \centering
            \label{fig:FFF6Dsimp}
            \includegraphics[scale=.75]{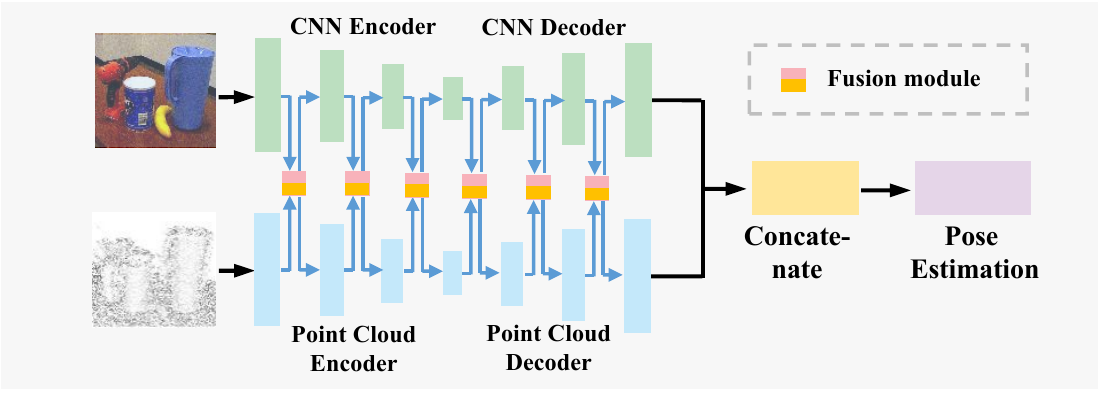}\\
        \end{minipage}%
    }%
    \caption{
        \textbf{Network Comparison} 
    }
    \label{fig:network_cmp}
\end{figure}

One line of existing works \cite{xiang2017posecnn,li2018unifiedMCN} leverage the advantage of the two data sources within cascaded designs. These works first estimate an initial pose from RGB images and then refine it on point clouds using either the Iterative Closest Point (ICP) algorithm or multi-view hypothesis verification. Such refinement procedures are time-consuming and can not be optimized with the pose from RGB images end-to-end.  On the other hand, works like \cite{qi2018frustum, xu2018pointfusion} apply a point cloud network (PCN) and a CNN to extract dense features from the cropped RGB image and point cloud respectively and the extracted dense features are then concatenated for pose estimation \cite{wang2019densefusion}. Recently, DenseFusion \cite{wang2019densefusion} proposed a better fusion strategy which replaced the naive concatenation operation with a dense fusion module, shown in Figure \ref{fig:dense_fusion}, and delivered improved performance.  
However, both feature concatenation and DenseFusion suffers from performance degeneration due to the separation of CNN and PCN in several scenarios, including objects with similar appearance or with reflective surfaces. Such cases are challenging either for the isolated CNN or the PCN feature extraction. 

In this work, we propose a \textit{full flow bidirectional} fusion network that perform fusion on each encoding and decoding layers for representation learning from the RGBD image, shown in Figure \ref{fig:FFF6Dsimp}. Our key insight is that appearance information in RGB and geometry information in point cloud can serve as complementary information during their feature extraction procedure. Specifically, during the CNN encoding-decoding procedure, it's hard for CNN to learn a distinctive representation for similar objects from the RGB image, which, however, is obvious in the PCN's view. On the other hand, the miss of depth caused by reflective surfaces of objects challenges the point cloud only geometry reasoning. Whereas, these objects are visible by CNN from RGB images. Hence, it's necessary to get through the two separated feature extraction branches in the early encoding-decoding stages and the proposed \textit{full flow bidirectional} fusion mechanism bridges this information gap. 

We further leverage the learned rich appearance and geometry representation for the pose estimation stage. We follow the pipeline proposed in PVN3D \cite{he2020pvn3d}, which opens up new opportunities for the 3D keypoint based 6D pose estimation. However, it only considers the distance between keypoints for 3D keypoint selection. Some selected keypoints might appear in non-salient regions like smooth surfaces without distinctive texture, making it hard to locate. Instead, we take both the object texture and geometry information into account and propose the SIFT-FPS algorithm for automatic 3D keypoint selections. Salient keypoints filtered in this way are easier for the network to locate and the pose estimation performance is facilitated.

To fully evaluate our method, we conduct experiments on three popular benchmark datasets, the YCB-Video, LineMOD, and Occlusion LineMOD datasets. Experimental results show that the proposed approach without any time-consuming post-refinement procedure outperforms the state-of-the-art by a large margin. 

To summarize, the main contributions of this work are:
\begin{itemize}
\item A novel full flow bidirectional fusion network for representation learning from a single scene RGBD image, which can be generalized to more applications, such as 3D object detection. 
\item A simple but effective 3D keypoint selection algorithm that leverages texture and geometry information of object models.
\item State-of-the-art 6D pose estimation performance on the YCB-Video, LineMOD, and Occlusion LineMOD datasets.
\item In-depth analysis to understand various design choices of the system.
\end{itemize}
\section{Related Work}
\subsection{Pose Estimation with RGB Data}
This line of works can be divided into three classes, holistic approaches, dense correspondence exploring, and 2D-keypoint-based. Holistic approaches \cite{huttenlocher1993comparing,gu2010discriminative,hinterstoisser2011gradient,xiang2017posecnn,li2018deepim,tulsiani2015viewpoints,su2015render,sundermeyer2018implicit,park2020latentfusion,shao2020pfrl} directly output pose parameters from RGB images. 
The non-linearity of rotation space limit the generalization of these approaches. Instead, dense correspondence approaches \cite{liebelt2008independent,sun2010depth,glasner2011aware,brachmann2014learning,michel2017global,kehl2016deep,doumanoglou2016recovering,li2019cdpn,wang2019normalized,chen2020learning,chen2020learning,hodan2020epos,cai2020reconstruct} find the correspondence between image pixels and mesh vertexes and recover poses within Perspective-n-Point (PnP) manners. Though robust to occlusion, the large output space limits the prediction accuracy. Instead, 2D-keypoint-based \cite{rublee2011orb,rothganger20063d,newell2016stacked,kendall2015posenet,oberweger2018making,peng2019pvnet,zhao2020learning3Dkp6DPose,liu2020keypose} detect 2D keypoints of objects to build the 2D-3D correspondence for pose estimation. However, the loss of geometry information due to perspective projections limit the performance of these RGB only methods.

\subsection{Pose Estimation with Point Clouds}
The development of depth sensors and point cloud representation learning techniques \cite{qi2017pointnet,qi2017pointnet++} motivates several point clouds only approaches. 
These approaches either utilize 3D ConvNets \cite{song2014sliding,song2016deep} or point cloud network \cite{zhou2018voxelnet,qi2019deep,hu2020randla} for feature extraction and 3D bounding box prediction. 
However, sparsity and non-texture of point cloud limit the performance of these approaches. Besides, objects with reflective surfaces can not be captured by depth sensors. Therefore, taking RGB images into account is necessary.

\newcommand{\mnps}{1}
\begin{figure*}
    \centering
    \subfigure[\textbf{The pipeline of FFB6D}. A CNN and a point cloud network is utilized for representation learning of RGB image and point cloud respectively. In flow of the two networks, bidirectional fusion modules are added as communicate bridges. The extracted per-point features are then fed into an instance semantic segmentation and a 3D keypoint voting modules to obtain per-object 3D keypoints. Finally, the pose is recovered within a least-squares fitting algorithm.]{
        \begin{minipage}[]{\mnps\linewidth}
            \centering
            \label{fig:network_full}
            \includegraphics[scale=.52]{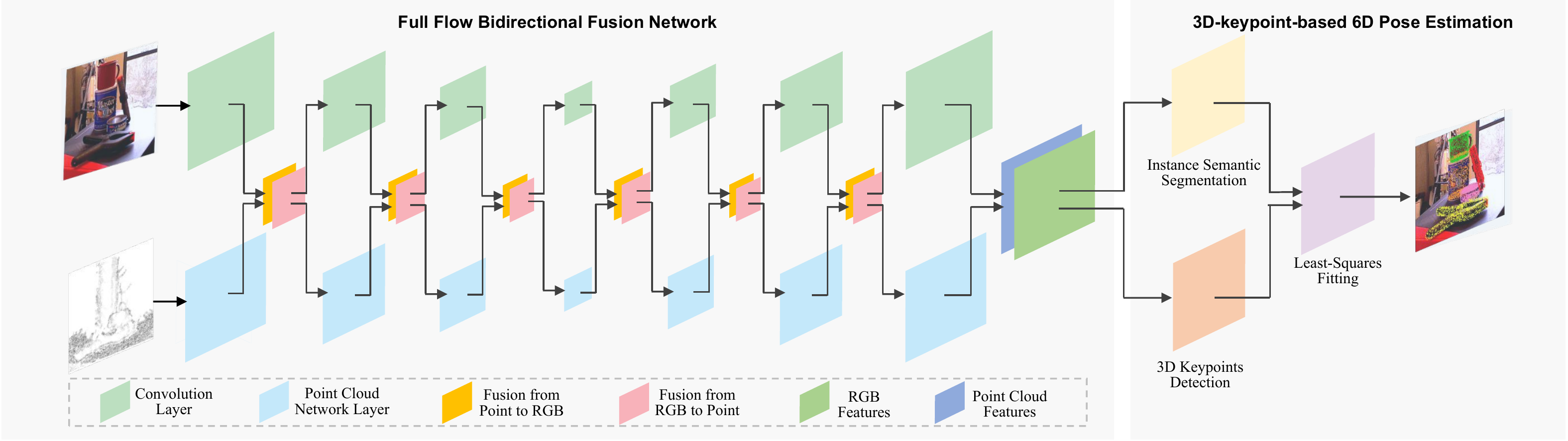}\\
        \end{minipage}%
    }%
    
    \subfigure[\textbf{Dense bidirectional fusion modules}. (1) The pixel-to-point fusion modules fuse RGB features to point cloud features. For each point, we find its $K_{r2p}$ nearest neighbors in the XYZ map and gather their corresponding appearance features from the RGB feature map. These features are then processed by max pooling and a shared MLP to obtain the most significant appearance features. Finally, a shared MLP fuses the concatenation of the appearance and geometry features to obtain the fused point features. (2) The point-to-pixel fusion modules similarly obtain fused pixel features as the pixel-to-point fusion.]{
        \begin{minipage}[]{\mnps\linewidth}
            \centering
            \label{fig:network_fusion}
            \includegraphics[scale=.62]{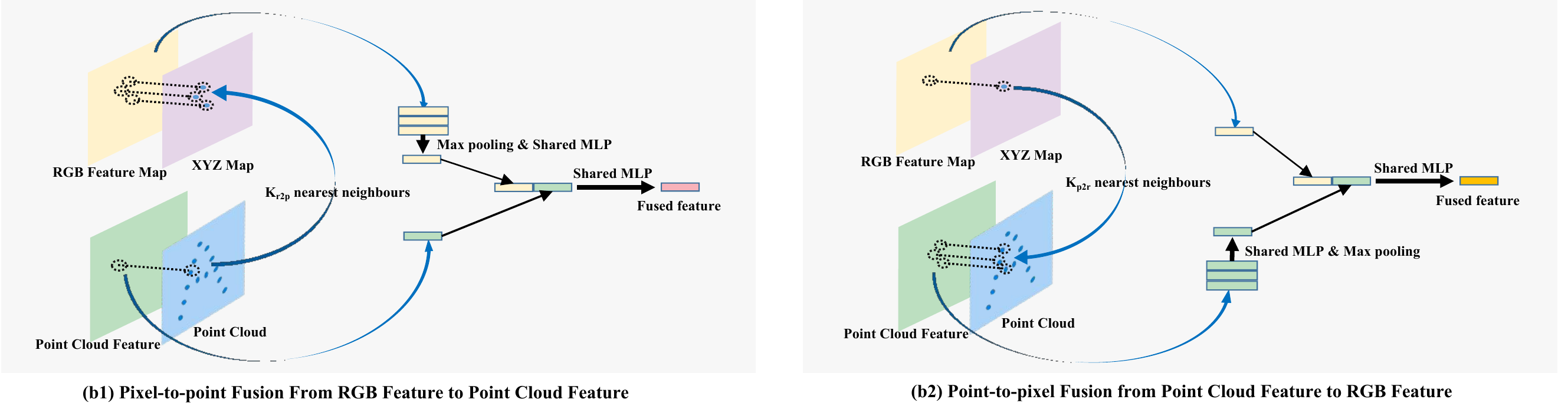}\\
        \end{minipage}%
    }%
    \caption{\textbf{Overview of FFB6D.}}
    \label{fig:network}
\end{figure*}

\subsection{Pose Estimation with RGB-D Data}
Traditional methods utilize hand-coded \cite{hinterstoisser2011multimodal,hinterstoisser2012model,rios2013discriminatively} templates or features optimized by surrogate objectives \cite{brachmann2014learning,tejani2014latent,wohlhart2015learning,kehl2016deep} from RGBD data and perform correspondence grouping and hypothesis verification. Recent data-driven approaches propose initial pose from RGB images and refine it with point cloud using ICP \cite{xiang2017posecnn} or MCN \cite{li2018unifiedMCN} algorithms. However, they are time-consuming and are not end-to-end optimizable. Instead, works like \cite{ku2018joint,liang2018DCFdeepcontiguous} add appearance information from CNN on RGB images as complementary information for geometry reasoning on bird-eye-view (BEV) images of point clouds. But they neglect the help of geometry information for RGB representation learning, the BEV ignore the pitch and roll of object pose, and the regular 2D CNN is not good at contiguous geometry reasoning either. Instead, \cite{xu2018pointfusion,qi2018frustum,wang2019densefusion,zhou2020novel,wada2020morefusion} extract features from RGB images and point clouds using CNN and point cloud network individually and then fused them for pose estimation. Such approaches are more effective and efficient. Nevertheless, since the appearance and geometry features are extracted separately, the two networks are not able to communicate and share information, and thus limit the expression ability of the learned representation. In this work, we add bidirectional fusion modules in the full network flow as communication bridges between the two networks. Assisted by supplementary information from another branch, better representation of appearance and geometry features are obtained for pose estimation.

\section{Proposed Method}
Given an RGBD image, the task of object 6D pose estimation aims to predict a transformation matrix that transforms the object from its coordinate system to the camera coordinate system. Such transformation consists of a rotation matrix $R \in SO(3)$ and a translation matrix $T \in \mathbb{R}^3$. To tackle the problem, pose estimation algorithms should fully explore the texture and the geometric information of both the scene and the target object. 

\subsection{Overview}
We propose a full flow bidirectional fusion network to solve the problem, as shown in Figure \ref{fig:network_full}. The proposed framework first extracts the pointwise RGBD feature for per-object 3D keypoints localization. Then the pose parameters are recovered within a least-square fitting manner. More specifically, given an RGBD image as input, we utilize a CNN to extract appearance features from the RGB image, and a point cloud network to extract geometric features from point clouds. During the feature extraction flow of the two networks, point-to-pixel and pixel-to-point fusion modules are added into each layer as communication bridges. In this way, the two branches can utilize the extra appearance (geometric) information from the other to facilitate their own representation learning. The extracted pointwise features are then fed into an instance semantic segmentation and a 3D keypoint detection module to obtain per-object 3D keypoints in the scene. Finally, a least-square fitting algorithm is applied to recover the 6D pose parameters. 

\subsection{Full Flow Bidirectional Fusion Network}
Given an aligned RGBD image, we first lift the depth image to a point cloud with the camera intrinsic matrix. A CNN and a point cloud network (PCN) is then applied for feature extraction from the RGB image and point cloud respectively. When the information flow through the two networks, point-to-pixel and pixel-to-point fusion modules are added for bidirectional communication. In this way, each branch can leverage local and global information from the other to facilitate their representation learning.

\textbf{Pixel-to-point fusion from image features to point cloud features.}
These modules share the appearance information extracted from CNN to the PCN. One naive way is to generate a global feature from the RGB feature map and then concatenate it to each point cloud feature. However, since most of the pixels are background and there are multi objects in the scene, squeezing the RGB feature map globally would lose lots of detailed information and harm the following pose estimation module. Instead, we introduce a novel pixel to point feature fusion module. Since the given RGBD image is well-aligned, we can use the 3D point clouds as a bridge to connect the pixel-wise and the point-wise features. More specifically, we lift the depth of each pixel to its corresponding 3D point with the camera intrinsic matrix and get an XYZ map, aligned with the RGB map. As shown in the left part of Figure \ref{fig:network_fusion}, for each point feature with its 3D point coordinate, we find its $K_{r2p}$ nearest point in the XYZ map and gather their corresponding appearance features from the RGB feature map. We then use max polling to integrate these neighboring appearance features following \cite{qi2017pointnet++}, and apply shared Multi-Layer Perceptrons (MLPs) to squeeze it to the same channel size as the point cloud feature:
 \begin{equation}
     \label{eqn:nnr2pRGB}
     \begin{split}
     F_{r2p} = & MLP(\max_{i=1}^{K_{r2p}}{F_{r_i}}) ,
     \end{split}
 \end{equation}
where $F_{r_i}$ is the $i_{th}$ nearest pixel of RGB feature and $F_{r2p}$ the integrated one.
We then concatenate the integrated appearance feature $F_{r2p}$ with the point feature $F_{point}$ and use shared MLP to obtained the fused point feature:
\begin{equation}
     \label{eqn:nnr2pPoint}
     \begin{split}
     F_{fused_p} = & MLP(F_{point} \oplus F_{r2p}) ,
     \end{split}
\end{equation}
where $\oplus$ is the concatenate operation.

One thing to mention is that in the flow of the appearance feature encoding, the height and width of the RGB feature maps get smaller when the network goes deeper. Therefore, we need to maintain a corresponding XYZ map so that each pixel of feature can find its 3D coordinate. Since the decreased size of the feature map is generated by convolution kernel scanning through the original feature map with stride, the centers of kernels become new coordinates of the feature maps. One simple way is to apply the same size of kernels to calculate the mean of XYZ within it to generate a new XYZ coordinate of a pixel. The corresponding XYZ map is then obtained by scanning through the XYZ map with the mean kernel in the same stride as CNN. However, noise points are produced by the mean operation as the depth changes remarkably on the boundary between the foreground objects and the background. Instead, a better solution is to resize the XYZ map to the same size as the feature map within the nearest interpolation algorithm.

\textbf{Point-to-pixel fusion from point cloud features to image features.}
These modules build bridges to transfer the geometric information obtained from the PCN to the CNN. The procedure is shown on the right side of Figure \ref{fig:network_fusion}. Same as the pixel-to-point fusion modules, we fuse the feature densely rather than naively concatenating the global point feature to each pixel. Specifically, for each pixel of feature with its XYZ coordinate, we find its $K_{p2r}$ nearest points from the point cloud and gather the corresponding point features. We squeeze the point features to the same channel size as the RGB feature and then use max pooling to integrate them. The integrated point feature is then concatenated to the corresponding color feature and mapped by a shared MLP to generate the fused one:
 \begin{equation}
     \label{eqn:nnr2pRGB}
     \begin{split}
     F_{p2r} = & \max_{j=1}^{K_{p2r}}{(MLP(F_{p_j})}) ,
     \end{split}
 \end{equation}
\begin{equation}
     \label{eqn:nnr2pPoint}
     \begin{split}
     F_{fused_r} = & MLP(F_{rgb} \oplus F_{p2r}) ,
     \end{split}
\end{equation}
where $F_{p_j}$ denotes the $j_{th}$ nearest point features, $F_{p2r}$ the integrated point features and $\oplus$ the concatenate operation.

\textbf{Dense RGBD feature embedding} 
With the proposed full flow fusion network, we obtain dense appearance embeddings from the CNN branch and dense geometry features from the PCN branch. We then find the correspondence between them by projecting each point to the image plane with the camera intrinsic matrix. According to the correspondence, we obtain pairs of appearance and geometry features and concatenate them together to form the extracted dense RGBD feature. These features are then fed into an instance semantic segmentation module and a 3D keypoint detection module for object pose estimation in the next step.

\subsection{3D Keypoint-based 6D Pose Estimation}
Recently, the PVN3D \cite{he2020pvn3d} work by He \textit{et al.} opens up new opportunities for using 3D keypoints to estimate object pose. In this work, we follow their 3D keypoint formulation but further improve the 3D keypoint selection algorithm to fully leverage the texture and geometry information of objects. Specifically, we first detect the per-object selected 3D keypoints in the scene and then utilize a least-squares fitting algorithm to recover the pose parameters.

\textbf{Per-object 3D keypoint detection} 
So far we have obtained the dense RGBD embeddings. We then follow PVN3D \cite{he2020pvn3d} and obtain the per-object 3D keypoints by adding an instance semantic segmentation module to distinguish different object instances and a keypoint voting module to recover 3D keypoints. The instance semantic segmentation module consists of a semantic segmentation module and a center point voting module, where the former one predicts per-point semantic labels, and the latter one learns the per-point offset to object centers for distinguishment of different instances. For each object instance, the keypoint voting module learns the point-wise offsets to the selected keypoints that vote for 3D keypoint within a MeanShift \cite{comaniciu2002mean} clustering manners.

\textbf{Keypoint selection} 
Previous works \cite{peng2019pvnet,he2020pvn3d} select keypoints from the target object surface using the Farthest Point Sampling (FPS) algorithm.  Specifically, they maintain a keypoint set initialized by a random point on the object surface and iteratively add other points that are farthest to those within the set until $N$ points are obtained. In this way, the selected keypoints spread on the object surface and stabilize the following pose estimation procedure \cite{peng2019pvnet,he2020pvn3d}. However, since the algorithm only takes the Euclidean distance into account, the selected points may appear in non-salient regions, such as flat planes without distinctive texture. These points are hard to detect and the accuracy of the estimated pose decrease. To fully leverage the texture and geometry information of objects, we propose a simple but effective 3D keypoint selection algorithm, named SIFT-FPS. Specifically, we use the SIFT \cite{lowe1999object} algorithm to detect 2D keypoints that are distinctive in texture images and then lift them to 3D. The FPS algorithm is then applied for the selection of top $N$ keypoints among them. In this way, the selected keypoints not only distribute evenly on the object surface but are also distinctive in texture and easy to detect.

\textbf{Least-Squares Fitting} Given the selected 3D keypoints in the object coordinates system $\{p_i\}_{i=1}^N$, and the corresponding 3D keypoints in the camera coordinated system $\{p^*_i\}_{i=1}^N$. The Least-Squares Fitting \cite{arun1987least} algorithm calculate the pose parameters $R$ and $T$ by minimizing the squared loss:
\begin{equation} \label{eqn:lsf}
    L_{\textrm{lsf}} = \sum_{i=1}^{N} || p^*_i - (R \cdot p_i + T) ||^2 .
\end{equation}

\renewcommand{\arraystretch}{1.5}
\newcommand{\ycbC}{0.7}
\begin{table*}[tp]
    \centering
    \fontsize{7.2}{7.5}\selectfont
    \begin{tabular}{l|C{\ycbC cm}|C{\ycbC cm}|C{\ycbC cm}|C{\ycbC cm}|C{\ycbC cm}|C{\ycbC cm}|C{\ycbC cm}|C{\ycbC cm}|C{\ycbC cm}|C{\ycbC cm}|C{\ycbC cm}|C{\ycbC cm}}
    \hline
                                  & \multicolumn{2}{c|}{PoseCNN \cite{xiang2017posecnn}}  & \multicolumn{2}{c|}{PointFusion \cite{xu2018pointfusion}}  & \multicolumn{2}{c|}{DCF \cite{liang2018DCFdeepcontiguous}} & \multicolumn{2}{c|}{DF (per-pixel) \cite{wang2019densefusion}}   & \multicolumn{2}{c|}{PVN3D \cite{he2020pvn3d}} & \multicolumn{2}{c}{Our FFB6D} \cr\hline
        Object                         & ADDS        & ADD(S)        & ADDS          & ADD(S)          & ADDs      & ADD(S)      & ADDS           & ADD(S)           & ADDS          & ADD(S)        & ADDS          & ADD(S)        \cr\hline
        002 master chef can            & 83.9        & 50.2          & 90.9          & -               & 90.9      & 74.6        & 95.3           & 70.7             & 96.0          & 80.5          & \textbf{96.3} & \textbf{80.6} \\
        003 cracker box                & 76.9        & 53.1          & 80.5          & -               & 87.1      & 79.3        & 92.5           & 86.9             & 96.1          & \textbf{94.8} & \textbf{96.3} & 94.6          \\
        004 sugar box                  & 84.2        & 68.4          & 90.4          & -               & 94.3      & 84.2        & 95.1           & 90.8             & 97.4          & 96.3          & \textbf{97.6} & \textbf{96.6} \\
        005 tomato soup can            & 81.0        & 66.2          & 91.9          & -               & 90.5      & 79.8        & 93.8           & 84.7             & \textbf{96.2} & 88.5          & 95.6          & \textbf{89.6} \\
        006 mustard bottle             & 90.4        & 81.0          & 88.5          & -               & 90.6      & 83.5        & 95.8           & 90.9             & 97.5          & 96.2          & \textbf{97.8} & \textbf{97.0} \\
        007 tuna fish can              & 88.0        & 70.7          & 93.8          & -               & 91.7      & 73.8        & 95.7           & 79.6             & 96.0          & \textbf{89.3} & \textbf{96.8} & 88.9          \\
        008 pudding box                & 79.1        & 62.7          & 87.5          & -               & 89.3      & 84.1        & 94.3           & 89.3             & \textbf{97.1} & \textbf{95.7} & \textbf{97.1} & 94.6          \\
        009 gelatin box                & 87.2        & 75.2          & 95.0          & -               & 92.9      & 89.5        & 97.2           & 95.8             & 97.7          & 96.1          & \textbf{98.1} & \textbf{96.9} \\
        010 potted meat can            & 78.5        & 59.5          & 86.4          & -               & 83.2      & 74.6        & 89.3           & 79.6             & 93.3          & \textbf{88.6} & \textbf{94.7} & 88.1          \\
        011 banana                     & 86.0        & 72.3          & 84.7          & -               & 84.8      & 71.0        & 90.0           & 76.7             & 96.6          & 93.7          & \textbf{97.2} & \textbf{94.9} \\
        019 pitcher base               & 77.0        & 53.3          & 85.5          & -               & 89.5      & 80.3        & 93.6           & 87.1             & 97.4          & 96.5          & \textbf{97.6} & \textbf{96.9} \\
        021 bleach cleanser            & 71.6        & 50.3          & 81.0          & -               & 88.4      & 79.8        & 94.4           & 87.5             & 96.0          & 93.2          & \textbf{96.8} & \textbf{94.8} \\
        \textbf{024 bowl}              & 69.6        & 69.6          & 75.7          & 75.7            & 80.3      & 80.3        & 86.0           & 86.0             & 90.2          & 90.2          & \textbf{96.3} & \textbf{96.3} \\
        025 mug                        & 78.2        & 58.5          & 94.2          & -               & 90.7      & 76.6        & 95.3           & 83.8             & \textbf{97.6} & \textbf{95.4} & 97.3          & 94.2          \\
        035 power drill                & 72.7        & 55.3          & 71.5          & -               & 87.4      & 78.4        & 92.1           & 83.7             & 96.7          & 95.1          & \textbf{97.2} & \textbf{95.9} \\
        \textbf{036 wood block}        & 64.3        & 64.3          & 68.1          & 68.1            & 84.2      & 84.2        & 89.5           & 89.5             & 90.4          & 90.4          & \textbf{92.6} & \textbf{92.6} \\
        037 scissors                   & 56.9        & 35.8          & 76.7          & -               & 84.2      & 70.3        & 90.1           & 77.4             & 96.7          & 92.7          & \textbf{97.7} & \textbf{95.7} \\
        040 large marker               & 71.7        & 58.3          & 87.9          & -               & 89.5      & 81.0        & 95.1           & 89.1             & \textbf{96.7} & \textbf{91.8} & 96.6          & 89.1          \\
        \textbf{051 large clamp}       & 50.2        & 50.2          & 65.9          & 65.9            & 63.6      & 63.6        & 71.5           & 71.5             & 93.6          & 93.6          & \textbf{96.8} & \textbf{96.8} \\
        \textbf{052 extra large clamp} & 44.1        & 44.1          & 60.4          & 60.4            & 64.4      & 64.4        & 70.2           & 70.2             & 88.4          & 88.4          & \textbf{96.0} & \textbf{96.0} \\
        \textbf{061 foam brick}        & 88.0        & 88.0          & 91.8          & 91.8            & 83.1      & 83.1        & 92.2           & 92.2             & 96.8          & 96.8          & \textbf{97.3} & \textbf{97.3} \cr\hline
        ALL                            & 75.8        & 59.9          & 83.9          & -               & 85.7      & 77.9        & 91.2           & 82.9             & 95.5          & 91.8          & \textbf{96.6} & \textbf{92.7} \cr\hline
    \end{tabular}
    \caption{Quantitative evaluation of 6D Pose without iterative refinement on the YCB-Video Dataset. The ADD-S \cite{xiang2017posecnn} and ADD(S) \cite{hinterstoisser2012model} AUC are reported. Symmetric objects are in bold. DF (per-pixel) means DenseFusion (per-pixel).
    }
    \label{tab:YCB_PFM}
\end{table*}

\newcommand{\kpC}{1.0}
\begin{table}[tp]
    \centering
    \fontsize{6.9}{6.8}\selectfont
    \begin{tabular}{L{0.9 cm}|C{\kpC cm}|C{0.9 cm}|C{\kpC cm}|C{1.1 cm}|C{\kpC cm}}
        \hline
                & PCNN+ICP & DF(iter.) & MoreFusion & PVN3D+ICP & FFB6D+ICP     \cr\hline
        ADD-S   & 93.0     & 93.2      & 95.7       & 96.1      & \textbf{97.0} \\
        ADD(S)  & 85.4     & 86.1      & 91.0       & 92.3      & \textbf{93.1} \\
        \hline  
    \end{tabular}
    \caption{Quantitative evaluation of 6D Pose with iterative refinement on the YCB-Video Dataset (ADD-S \cite{xiang2017posecnn} and ADD(S) AUC \cite{hinterstoisser2012model}). Baselines: PoseCNN+ICP \cite{xiang2017posecnn}, DF(iter.) \cite{wang2019densefusion}, MoreFusion \cite{wada2020morefusion}, PVN3D+ICP \cite{he2020pvn3d}.}
    \label{tab:YCB_PF_ICP}
\end{table}

\section{Experiments}
\subsection{Benchmark Datasets}
We evaluate our method on three benchmark datasets. 

\textbf{YCB-Video} \cite{calli2015ycb} contains 92 RGBD videos that capture scenes of 21 selected YCB objects. We followed previous works \cite{xiang2017posecnn,wang2019densefusion,he2020pvn3d} to split the training and testing set. Synthetic images were also taken for training as in \cite{xiang2017posecnn} and the hole completion algorithm \cite{ku2018defense} is applied for hole filling for depth images as in \cite{he2020pvn3d}. 

\textbf{LineMOD} \cite{hinterstoisser2011multimodal} is a dataset with 13 videos of 13 low-textured objects. The texture-less objects, cluttered scenes, and varying lighting make this dataset challenge. We split the training and testing set following previous works \cite{xiang2017posecnn,peng2019pvnet} and generate synthesis images for training following \cite{peng2019pvnet,he2020pvn3d}.

\textbf{Occlusion LINEMOD} \cite{OcclusionLMbrachmann2014learning} was selected and annotated from the LineMOD datasets. Each scene in this dataset consists of multi annotated objects, which are heavily occluded. The heavily occluded objects make this dataset challenge.

\begin{figure}
  \centering
  \includegraphics[scale=0.55]{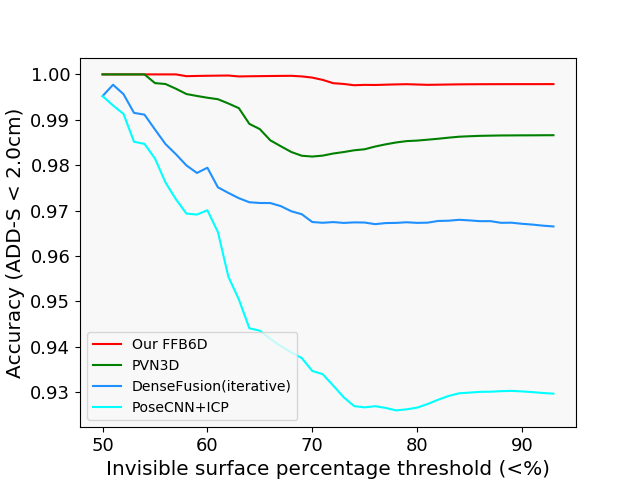}
  \caption{
    Performance of different approaches under increasing levels of occlusion on the YCB-Video dataset.
  }
  \label{fig:occlussion_auc}
\end{figure}

\subsection{Evaluation Metrics}
We use the average distance metrics ADD and ADD-S for evaluation. For asymmetric objects, the ADD metric calculate the point-pair average distance between objects vertexes transformed by the predicted and the ground truth pose, defined as follows:
\begin{equation}
    \label{eqn:ADD}
    \textrm{ADD} = \frac{1}{m} \sum_{v \in \mathcal{O}} || (Rv+T) - (R^*v + T^*) || .
\end{equation}
where $v$ denotes a vertex in object $\mathcal{O}$, $R$, $T$ the predicted pose and $R^*, T^*$ the ground truth.
For symmetric objects, the ADD-S based on the closest point distance is applied:
\begin{equation}
    \label{eqn:ADDS}
    \textrm{ADD{-}S} = \frac{1}{m} \sum_{v_1 \in \mathcal{O}} \min_{v_2 \in \mathcal{O}}{|| (Rv_1+T) - (R^*v_2 + T^*) ||} .
\end{equation}
In the YCB-Video dataset, we follows previous methods \cite{xiang2017posecnn,wang2019densefusion,he2020pvn3d} and report the area under the accuracy-threshold curve obtained by varying the distance threshold (ADD-S and ADD(S) AUC). In the LineMOD and Occlusion LineMOD datasets, we report the accuracy of distance less than 10\% of the objects diameter (ADD-0.1d) as in \cite{hinterstoisser2012model,peng2019pvnet}.

\subsection{Training and Implementation}
\textbf{Network architecture.} We apply ImageNet \cite{deng2009imagenet} pre-trained ResNet34 \cite{resnet} as encoder of RGB images, followed by a PSPNet \cite{zhao2017pyramid} as decoder.  For point cloud feature extraction, we randomly sample 12288 points from depth images following \cite{he2020pvn3d} and applied RandLA-Net \cite{hu2020randla} for representation learning. In each encoding and decoding layers of the two networks, max pooling and shared MLPs are applied to build bidirectional fusion modules. After the process of the full flow bidirectional fusion network, each point has a feature $f_i \in \mathbb{R}^C$ of $C$ dimension. These dense RGBD features are then fed into the instance semantic segmentation and the keypoint offset learning modules consist of shared MLPs.

\textbf{Optimization regularization.} The semantic segmentation branch is supervised by Focal Loss \cite{lin2017focal}. The center point voting and 3D keypoints voting modules are optimized by L1 loss as in \cite{he2020pvn3d}. To jointly optimize the three tasks, a multi-task loss with the weighted sum of them is applied following \cite{he2020pvn3d}.

\textbf{SIFT-FPS keypoint selection algorithm.} We put the target object at the center of a sphere and sample viewpoints of the camera on the sphere equidistantly. RGBD images with camera poses are obtained by render engines. We then detect 2D keypoints from RGB images with SIFT. These 2D keypoints are lifted to 3D and transformed back to the object coordinates system. Finally, an FPS algorithm is applied to select $N$ target keypoints out of them.

\renewcommand{\arraystretch}{1.3}
\begin{table*}[tp]
    \centering
    \fontsize{7.0}{6.8}\selectfont
    \begin{tabular}{l|C{1.1cm}|C{1.1cm}|C{1.1cm}|C{1.1cm}|C{1.1cm}|C{1.1cm}|C{1.1cm}|C{1.1cm}|C{1.1cm} }
        \hline
                & \multicolumn{4}{c|}{RGB}               & \multicolumn{5}{c}{RGB-D}                    \cr\hline                                               
                & PoseCNN DeepIM \cite{xiang2017posecnn,li2018deepim} & PVNet\cite{peng2019pvnet} & CDPN\cite{li2019cdpn} & DPOD\cite{Zakharov2019dpod}  & Point- Fusion\cite{xu2018pointfusion} & Dense- Fusion\cite{wang2019densefusion} & G2L-Net\cite{chen2020g2l} & PVN3D\cite{he2020pvn3d} & Our FFB6D          \cr\hline

MEAN            & 88.6           & 86.3  & 89.9 & 95.2  & 73.7        & 94.3                   & 98.7           & 99.4                      & \textbf{99.7} 
        \cr\hline  
    \end{tabular}
    \caption{Quantitative evaluation of 6D pose on the LineMOD dataset (ADD-0.1d \cite{hinterstoisser2012model} metrics).}
    \label{tab:LM_PFM}
\end{table*}

\newcommand{\OlC}{0.95}
\begin{table}[tp]
    \centering
    \fontsize{6.9}{6.8}\selectfont
    \begin{tabular}{l|C{\OlC cm}|C{\OlC cm}|C{\OlC cm}|C{\OlC cm}|C{\OlC cm} }
        \hline
        Method   & PoseCNN \cite{xiang2017posecnn} & Oberweger \cite{oberweger2018making} & Hu et al. \cite{hu2019segmentation}         & Pix2Pose \cite{park2019pix2pose} & PVNet \cite{peng2019pvnet}          \cr\hline
        ADD-0.1d & 24.9    & 27.0      & 27.0       & 32.0     & 40.8  \cr\hline
        Method   & DPOD \cite{Zakharov2019dpod}    & Hu et al.\cite{hu2020single} & HybridPose \cite{song2020hybridpose} & PVN3D \cite{he2020pvn3d}    & Our FFB6D \cr\hline
        ADD-0.1d & 47.3    & 43.3      & 47.5       & 63.2     & \textbf{66.2}  \\
        \hline  
    \end{tabular}
    \caption{Quantitative evaluation of 6D pose (ADD-0.1d) on the Occlusion-LineMOD dataset.}
    \label{tab:OCC_LM_PFM}
\end{table}

\subsection{Evaluation on Three Benchmark Datasets.}
We evaluate the proposed models on the YCB-Video, the LineMOD, and the Occlusion LineMOD datasets.

\textbf{Evaluation on the YCB-Video dataset.} Table \ref{tab:YCB_PFM} shows the quantitative evaluation results of the proposed FFB6D on the YCB-Video dataset. We compare it with other single view methods without iterative refinement. FFB6D advances state-of-the-art results by 1.1\% on the ADD-S metric and 1.0\% on the ADD(S) metric. Equipped with extra iterative refinement, our approach also achieves the best performance, demonstrated in Table \ref{tab:YCB_PF_ICP}. Note that the proposed FFB6D without any iterative refinement even outperforms state-of-the-arts that require time-consuming post-refinement procedures. Qualitative results are reported in the supplementary material. 

\textbf{Robustness towards occlusion.} We follow \cite{wang2019densefusion,he2020pvn3d} to report the ADD-S less than 2cm accuracy under the growth of occlusion level on the YCB-Video dataset. As is shown in Figure \ref{fig:occlussion_auc}, previous methods degrade as the occlusion increase. In contrast, FFB6D didn't suffer from a drop in performance. We think our full flow bidirectional fusion mechanism makes full use of the texture and geometry information in the captured data and enables our approach to locate 3D keypoints even in highly occluded scenes. 

\textbf{Evaluation on the LineMOD dataset \& Occlusion LineMOD dataset.} The proposed FFB6D outperforms the state-of-the-art on the LineMOD dataset, presented in Table \ref{tab:LM_PFM}. We also evaluate FFB6D on the Occlusion LineMOD dataset, shown in Table \ref{tab:OCC_LM_PFM}. In the table, our FFB6D without iterative refinement advances state-of-the-art by 4.7\%, further confirming its robustness towards occlusion.

\begin{figure*}
  \centering
     \includegraphics[scale=0.55]{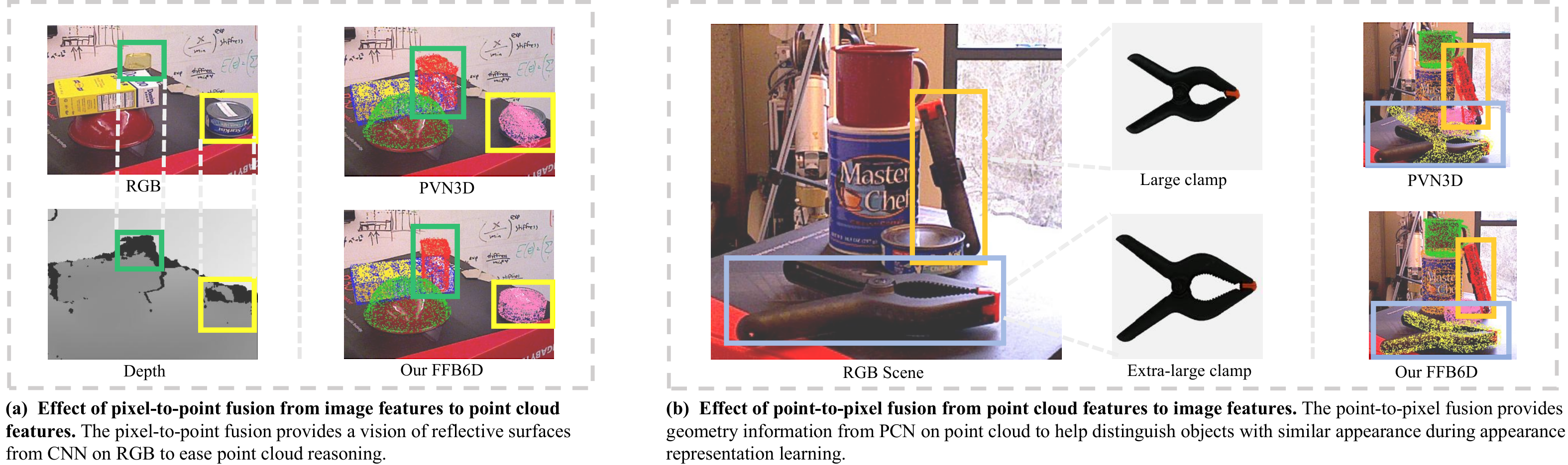}
     \caption{
         Effect of full flow bidirectional fusion, compared to PVN3D \cite{he2020pvn3d} with DenseFusion architecture.
     }
     \label{fig:eff_BiF}
\end{figure*}

\subsection{Ablation Study}
In this subsection, we present extensive ablation studies on our design choices and discuss their effect. 

\newcommand{\fdeC}{1.15}
\begin{table}[tp]
  \centering
  \fontsize{6.9}{6.8}\selectfont
  \begin{tabular}{C{\fdeC cm}|C{\fdeC cm}|C{\fdeC cm}|C{\fdeC cm}|C{\fdeC cm} }
    \hline
    \multicolumn{3}{c|}{Fusion Stage} & \multicolumn{2}{c}{Pose Result} \cr\hline
    FE     & FD    & DF    & ADD-S    & ADD(S) \cr\hline
          &     &     & 91.9     & 87.5 \cr\hline
    $\surd$  &     &     & 96.2     & 92.2 \cr\hline
          & $\surd$ &     & 93.0     & 89.2 \cr\hline
          &     & $\surd$ & 94.0     & 90.8 \cr\hline
    $\surd$  & $\surd$ &     & \textbf{96.6} & \textbf{92.7} \cr\hline
    $\surd$  & $\surd$ & $\surd$ & 96.4     & 92.6  
    \cr\hline 
  \end{tabular}
  \caption{Effect of fusion stages on the YCB-Video dataset. FE: fusion during encoding; FD: fusion during decoding; DF: Dense Fusion on the two final feature maps.}
  \label{tab:EffFuseEcDc}
\end{table}

\newcommand{\fprC}{1.15}
\begin{table}[tp]
  \centering
  \fontsize{6.9}{6.8}\selectfont
  \begin{tabular}{C{\fprC cm}|C{\fprC cm}|C{\fprC cm}|C{\fprC cm} }
    \hline
    \multicolumn{2}{c|}{Fusion Direction} & \multicolumn{2}{c}{Pose Result} \cr\hline
    P2R    & R2P   & ADD-S    & ADD(S)  \cr\hline
          &     & 91.9     & 87.5   \cr\hline
    $\surd$  &     & 95.8     & 91.1   \cr\hline
          & $\surd$ & 94.5     & 90.6   \cr\hline
    $\surd$  & $\surd$ & \textbf{96.6} & \textbf{92.7}
    \cr\hline
  \end{tabular}
  \caption{Effect of fusion direction on the YCB-Video dataset. P2R means fusion from point cloud embeddings to RGB embeddings, and R2P means fusion from RGB embeddings to point cloud embeddings.}
  \label{tab:EffFuseRGBPoint}
\end{table}

\textbf{Effect of full flow bidirectional fusion.}
To validate that building fusion modules between the two modality networks in full flow help, we ablate fusion stages in Table \ref{tab:EffFuseEcDc}. Compared to the mechanism without fusion, adding fusion modules either on the encoding stage, decoding stage, or on the final feature maps, can all boost the performance. Among the three stages, fusion on the encoding stages obtained the highest improvement. We think that's because the extracted local texture and geometric information are shared through the fusion bridge on the early encoding stage, and more global features are shared when the network goes deeper. Also, adding fusion modules in full flow of the network, saying that on both the encoding and decoding stages, obtains the highest performance. While adding extra DenseFusion behind full flow fusion obtains no performance gain as the two embeddings have been fully fused.

We also ablate the fusion direction in Table \ref{tab:EffFuseRGBPoint} to validate the help of bidirectional fusion. Compared with no fusion, both the fusion from RGB to point cloud and the inverse way facilitate better representation learning. Combining the two obtains the best results. On the one hand, from the view of PCN, we think the rich textures information obtained from high-resolution RGB images helps semantic recognition. In addition, the high-resolution RGB features provide rich information for blind regions of depth sensors caused by reflective surfaces. It serves as a completion to point cloud and improve pose accuracy, as shown in Figure \ref{fig:eff_BiF}(a). On the one hand, geometry information extracted from point cloud helps the RGB branch by distinguishing foreground objects from the background that are in similar colors. Moreover, the shape size information extracted from point clouds helps divide objects with a similar appearance but in a different size, as is shown in Figure \ref{fig:eff_BiF}(b).

\textbf{Effect of representation learning frameworks.} We explore the effect of different representation learning frameworks for the two modalities of data in this part. The result is presented in Table \ref{tab:EffBackB}. We find that neither concatenating the XYZ map as extra information to CNN (CNN-R$\oplus$D) nor adding RGB values as extra inputs to the PCN (PCN-R$\oplus$D) achieves satisfactory performance. Using two CNNs (CNN-R+CNN-D) or two PCNs (PCN-R+PCN-D) with full flow bidirectional fusion modules get better but are still far from satisfactory. In contrast, applying CNN on the RGB image and PCN on the point cloud (CNN-R+PCN-D) gets the best performance. We think that the grid-like image data is discrete, on which the regular convolution kernel fits better than continuous PCN. While the geometric information residing in the depth map is defined in a continuous vector space, and thus PCNs can learn better representation.

\newcommand{\bbC}{1.7}
\begin{table}[tp]
  \centering
  \fontsize{6.9}{6.8}\selectfont
  \begin{tabular}{l|C{\bbC cm}|C{\bbC cm}|C{\bbC cm} }
    \hline
           & CNN-R$\oplus$D    & PCN-R$\oplus$D    & CNN-R+CNN-D \cr\hline
    ADD-S  & 91.0        & 90.9        & 92.1          \\
    ADD(S) & 85.5        & 81.2        & 87.2          \cr\hline
           & PCN-R+PCN-D & CNN-R+3DC-D & CNN-R+PCN-D \cr\hline
    ADD-S  & 91.8        & 94.9        & \textbf{96.6} \\
    ADD(S) & 84.3        & 90.1        & \textbf{92.7}
    \cr\hline 
  \end{tabular}
  \caption{Effect of representation learning framework on the two modalities of data. CNN: 2D Convolution Neural Network; PCN: point cloud network; 3DC: 3D ConvNet; R: RGB images; D: XYZ maps for CNN, point clouds for PCN and voxelized point clouds for 3D ConvNet.}
  \label{tab:EffBackB}
\end{table}

\textbf{Effect of 3D keypoints selection algorithm.} In Table \ref{tab:EffKPSA}, we study the effect of different keypoint selection algorithms. Compared with FPS that only considers the mutual distance between keypoints, our SIFT-FPS algorithm taking both object texture and geometry information into account is easier to locate. Therefore, the predicted keypoint error is smaller and the estimated poses are more accurate. 

\newcommand{\cmpkpC}{0.62}
\begin{table}[tp]
  \centering
  \fontsize{6.9}{6.8}\selectfont
  \begin{tabular}{l|C{\cmpkpC cm}|C{\cmpkpC cm}|C{\cmpkpC cm}|C{\cmpkpC cm}|C{\cmpkpC cm}|C{\cmpkpC cm} }
    \hline
             & FPS4 & S-F4 & FPS8 & S-F8          & FPS12          & S-F12 \cr\hline
KP err. (cm) & 1.3   & 1.2   & 1.4   & \textbf{1.2}   & 1.6             & 1.3   \\
ADD-S        & 95.9  & 96.4  & 96.1  & \textbf{96.6}  & 96.0            & 96.5  \\
ADD(S)       & 92.0  & 92.4  & 92.3  & \textbf{92.7}  & 92.0            & 92.6 
    \cr\hline 
  \end{tabular}
  \caption{Effect of keypoint selection algorithm. S-F means the proposed SIFT-FPS algorithm.}
  \label{tab:EffKPSA}
\end{table}

\textbf{Effect of the downsample strategy of the assisting XYZ map.} The size of RGB feature maps are shrunk by stridden convolution kernels. To maintain the corresponding XYZ maps, we first scale it down with the same size of mean kernels and got 96.3 ADD-S AUC. However, simply resize the XYZ map with the nearest interpolation got 96.6. We find the average operation produces noise points on the boundary and decrease the performance. 

\textbf{Model parameters and time efficiency.}
In Table \ref{tab:ParamEff}, we report the parameters and run-time breakdown of FFB6D. Compared to PVN3D \cite{he2020pvn3d}, which obtained the fused RGBD feature by dense fusion modules \cite{wang2019densefusion} in the final layers, our full flow bidirectional fusion network achieve better performance with fewer parameters and is 2.5 times faster.

\newcommand{\ptC}{1.15}
\begin{table}[tp]
  \centering
  \fontsize{6.9}{6.8}\selectfont
  \begin{tabular}{l|C{\ptC cm}|C{\ptC cm}|C{\ptC cm}|C{\ptC cm} }
    \hline
    \multicolumn{1}{l|}{}         & \multicolumn{1}{c|}{}              & \multicolumn{3}{c}{Run-time (ms/frame)}                   \\ \cline{3-5}
    \multicolumn{1}{l|}{\multirow{-2}{*}{}} & \multicolumn{1}{c|}{\multirow{-2}{*}{Parameters}} & \multicolumn{1}{c|}{NF} & \multicolumn{1}{c|}{PE} & \multicolumn{1}{c}{All} \\ \hline
    PVN3D\cite{he2020pvn3d}         & 39.2M           & 170                 & 20                 & 190        \\
    Our FFB6D                & 33.8M           & 57                 & 18                 & 75 \cr
    \hline 
  \end{tabular}
  \caption{Model parameters and run-time breakdown on the LineMOD dataset. NF: Network Forward; PE: Pose Estimation. Our FFB6D with fewer parameters is 2.5x faster.
  }
  \label{tab:ParamEff}
\end{table}


\section{Conclusion}
We propose a novel full flow bidirectional fusion network for representation learning from a single RGBD image, which extract rich appearance and geometry information in the scene for pose estimation. Besides, we introduce a simple but effective SIFT-FPS keypoint selection algorithms that leverage texture and geometry information of objects to simplify keypoint localization for precise pose estimation. Our approach outperforms all previous approaches in several benchmark datasets by remarkable margins. Moreover, we believe the proposed full flow bidirectional fusion network can generalize to more applications built on RGBD images, such as 3D object detection, 3D instance semantic segmentation and salient object detection etc. and expect to see more future research along this line.

\appendix
\section{Appendix}
\subsection{Details of the Network Architecture}
\begin{figure*}
    \centering
    \includegraphics[scale=0.8]{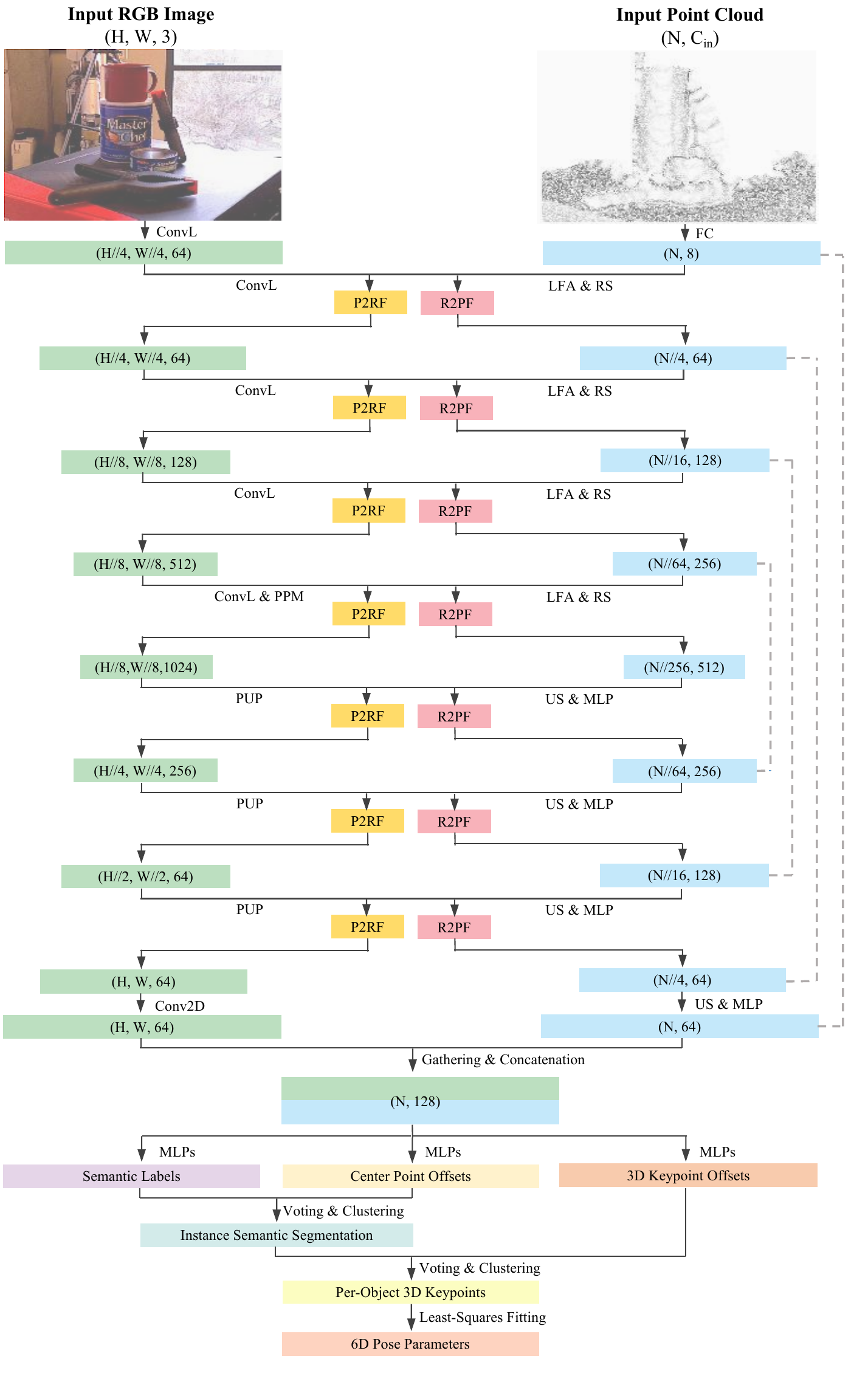}
    \caption{
        \textbf{The detailed architecture of our FFB6D.} For the convolution neural network (CNN) branch on the RGB image, we utilize ResNet34 \cite{resnet} and PSPNet \cite{zhao2017pyramid} as encoder and decoder. ConvL: Convolution Layers of ResNet34, PPM: Pyramid Pooling Modules of PSPNet, PUP: PSPNet Up-sampling, Conv2D: 2D convolution layer. For the point cloud network (PCN) branch on the point cloud, we apply RandLA-Net \cite{hu2020randla} for feature extraction. FC: Fully Connected layer, LFA: Local Feature Aggregation, RS: Random Sampling, MLP: shared Multi-Layer Perceptron, US: Up-sampling. In the flow of the two networks, point-to-pixel fusion modules, P2RF, and pixel-to-point fusion modules, R2PF consists of max pooling and shared MLPs are added. The extracted features from the two networks are then concatenated and fed into the following semantic segmentation, center point voting and 3D keypoints voting modules \cite{he2020pvn3d} composed by shared MLPs. A clustering algorithm is then applied to distinguish different instances with the same semantic labels and points on the same instance vote for their target keypoints. With detected 3D keypoints, a least-squares fitting algorithm is applied to recover the pose parameters.
    }
    \label{fig:FFB6D_detailed}
\end{figure*}

Figure \ref{fig:FFB6D_detailed} shows the detailed architecture of the proposed FFB6D. We applied ImageNet \cite{deng2009imagenet} pre-trained ResNet34 \cite{resnet} and PSPNet \cite{zhao2017pyramid} as encoder and decoder of the input RGB image. Meanwhile, a RandLA-Net \cite{hu2020randla} is applied for point cloud representation learning. On each encoding and decoding layer of the two networks, point-to-pixel and pixel-to-point fusion modules are added for information communication. Finally, the extracted dense appearance and geometry features are concatenated and fed into the semantic segmentation, center point voting, and 3D keypoints voting modules for pose estimation. Details of each part are as follows:

\textbf{Network Input.} 
The input of the convolution neural network (CNN) branch is a full scene image with a size of $H \times W \times 3$, where $H$ is the height of the RGB image, $W$ the width, and $3$ the three channels of color information (RGB). For the point cloud learning branch, the input is a randomly subsampled point cloud from the scene depth image, with a size of $N \times C_{in}$, where $N$ set to 12288 is the number of sampled points, and $C_{in}$ the input coordinate, color and normal information of each point (x-y-z-R-G-B-nx-ny-nz). 

\textbf{Encoding Layers.} 
We utilize ResNet34 \cite{resnet} as the encoder of RGB images, which consists of five convolution layers to reduce the size of feature maps and increase the number of feature channels. The Pyramid Pooling Modules (PPM) from PSPNet \cite{zhao2017pyramid} is also applied in the last encoding layer. Meanwhile, in the point cloud network branch, after being processed by a fully connected layer, the point features are fed into four encoding layers of RandLA-Net \cite{hu2020randla} for feature encoding, each of which consists of a local feature aggregation module and a random sampling operation designed in the work \cite{hu2020randla}.

\textbf{Decoding Layers.} 
In the decoding stage, three up-sampling modules and a final convolution layer from PSPNet are used for appearance feature decoding. Meanwhile, in the point cloud network branch, four decoding layers from RandLA-Net are utilized as point cloud features decoders, which consists of the random sampling operations and local feature aggregation modules designed in \cite{hu2020randla}.

\textbf{Bidirectional Fusion Modules.} 
On each encoding and decoding stage, point-to-pixel and pixel-to-point fusion modules (Section 3.2) are added for bidirectional information communication. For each pixel-to-point fusion module, we set $K_{r2p} = 16$ and aggregate 16 nearest pixel of appearance features through a max-pooling and a single layer shared MLP, $MLP[c_{r}, c_{p}]$, where $c_{r}$ denotes the channel size of RGB features and $c_{p}$ the channel size of corresponding point features. The aggregated pixels of appearance features are then concatenated with the corresponding point features and map by a shared MLP, $MLP[2*c_p, c_p]$ to generate each fused point feature. Meanwhile, we set $K_{p2r} = 1$ and get the fused appearance features similarly in each point-to-pixel fusion module.

\textbf{Prediction Headers.}
Three headers are added after the extracted dense RGBD features to predict the semantic label, center point offset as wel as the 3D keytpoints offsets of each point. These headers consists of shared MLPs, denoted as $MLP[c_r+c_p, c_1, c_2, ..., c_k]$, where $c_r$ and $c_p$ represent the channel size of extracted appearance and geometry features respectively, and $c_i$ the output channel size of the $i$-th layer in the MLP. Specifically, the semantic segmentation module consists of $MLP[c_r+c_p, 128, 128, 128, n_{cls}]$, the center offset learning module comprises $MLP[c_r+c_p, 128, 128, 128, 3]$,  and the 3D keypoints offset module is composed of $MLP[c_r+c_p, 128, 128, 128, n_{kps} * 3]$, where $n_{cls}$ denotes number of object classes and $n_{kps}$ means the number of keypoints of each object.

\textbf{Pose Estimation Modules.}
Given the predicted semantic label and center point offset of each point in the scene, a MeanShift \cite{comaniciu2002mean} clustering algorithm is applied to distinguish different object instances with the same semantic. Then, for each instance, each point within it votes for its 3D keypoint with the MeanShift \cite{comaniciu2002mean} algorithm. Finally, a least-squares fitting algorithm is applied to recover the object pose parameters according to the detected 3D keypoints.

\renewcommand{\arraystretch}{1.3}
\begin{table*}[tp]
    \centering
    \fontsize{7.0}{6.8}\selectfont
    \begin{tabular}{l|C{1.1cm}|C{1.1cm}|C{1.1cm}|C{1.1cm}|C{1.1cm}|C{1.1cm}|C{1.1cm}|C{1.1cm}|C{1.1cm} }
        \hline
                & \multicolumn{4}{c|}{RGB}               & \multicolumn{5}{c}{RGB-D}                    \cr\hline                                               
                & PoseCNN DeepIM \cite{xiang2017posecnn,li2018deepim} & PVNet\cite{peng2019pvnet} & CDPN\cite{li2019cdpn} & DPOD\cite{Zakharov2019dpod}  & Point- Fusion\cite{xu2018pointfusion} & Dense- Fusion\cite{wang2019densefusion} & G2L-Net\cite{chen2020g2l} & PVN3D\cite{he2020pvn3d} & Our FFB6D          \cr\hline
ape             & 77.0           & 43.6  & 64.4 & 87.7  & 70.4        & 92.3                   & 96.8           & 97.3                      & \textbf{98.4}  \\
benchvise       & 97.5           & 99.9  & 97.8 & 98.5  & 80.7        & 93.2                   & 96.1           & 99.7                      & \textbf{100.0} \\
camera          & 93.5           & 86.9  & 91.7 & 96.1  & 60.8        & 94.4                   & 98.2           & 99.6                      & \textbf{99.9}  \\
can             & 96.5           & 95.5  & 95.9 & 99.7  & 61.1        & 93.1                   & 98.0           & 99.5                      & \textbf{99.8}  \\
cat             & 82.1           & 79.3  & 83.8 & 94.7  & 79.1        & 96.5                   & 99.2           & 99.8                      & \textbf{99.9}  \\
driller         & 95.0           & 96.4  & 96.2 & 98.8  & 47.3        & 87.0                   & 99.8           & 99.3                      & \textbf{100.0} \\
duck            & 77.7           & 52.6  & 66.8 & 86.3  & 63.0        & 92.3                   & 97.7           & 98.2                      & \textbf{98.4}  \\
\textbf{eggbox} & 97.1           & 99.2  & 99.7 & 99.9  & 99.9        & 99.8                   & \textbf{100.0} & 99.8                      & \textbf{100.0} \\
\textbf{glue}   & 99.4           & 95.7  & 99.6 & 96.8  & 99.3        & \textbf{100.0}         & \textbf{100.0} & \textbf{100.0}            & \textbf{100.0} \\
holepuncher     & 52.8           & 82.0  & 85.8 & 86.9  & 71.8        & 92.1                   & 99.0           & \textbf{99.9}             & 99.8           \\
iron            & 98.3           & 98.9  & 97.9 & 100.0 & 83.2        & 97.0                   & 99.3           & 99.7                      & \textbf{99.9}  \\
lamp            & 97.5           & 99.3  & 97.9 & 96.8  & 62.3        & 95.3                   & 99.5           & 99.8                      & \textbf{99.9}  \\
phone           & 87.7           & 92.4  & 90.8 & 94.7  & 78.8        & 92.8                   & 98.9           & 99.5                      & \textbf{99.7}  \cr\hline
MEAN            & 88.6           & 86.3  & 89.9 & 95.2  & 73.7        & 94.3                   & 98.7           & 99.4                      & \textbf{99.7} 
        \cr\hline  
    \end{tabular}
    \caption{Quantitative evaluation on the LineMOD dataset. The ADD-0.1d \cite{hinterstoisser2012model} metric is reported and symmetric objects are in bold.}
    \label{tab:LM_PFM}
\end{table*}

\begin{table*}[tp]
    \centering
    \fontsize{7.0}{6.8}\selectfont
    \begin{tabular}{l|C{\OlC cm}|C{\OlC cm}|C{\OlC cm}|C{\OlC cm}|C{\OlC cm}|C{\OlC cm}|C{\OlC cm}|C{\OlC cm}|C{\OlC cm}|C{\OlC cm} }
        \hline
        Method   & PoseCNN \cite{xiang2017posecnn} & Oberweger \cite{oberweger2018making} & Hu et al. \cite{hu2019segmentation}         & Pix2Pose \cite{park2019pix2pose} & PVNet \cite{peng2019pvnet} & DPOD \cite{Zakharov2019dpod}    & Hu et al.\cite{hu2020single} & HybridPose \cite{song2020hybridpose} & PVN3D \cite{he2020pvn3d}    & Our FFB6D \cr\hline
        ape             & 9.6  & 12.1 & 17.6 & 22.0 & 15.8 & -    & 19.2 & 20.9 & 33.9          & \textbf{47.2}                     \\
        can             & 45.2 & 39.9 & 53.9 & 44.7 & 63.3 & -    & 65.1 & 75.3 & \textbf{88.6} & 85.2                              \\
        cat             & 0.9  & 8.2  & 3.3  & 22.7 & 16.7 & -    & 18.9 & 24.9 & 39.1          & \textbf{45.7}                     \\
        driller         & 41.4 & 45.2 & 62.4 & 44.7 & 65.7 & -    & 69.0 & 70.2 & 78.4          & \textbf{81.4}                     \\
        duck            & 19.6 & 17.2 & 19.2 & 15.0 & 25.2 & -    & 25.3 & 27.9 & 41.9          & \textbf{53.9}                     \\
        \textbf{eggbox} & 22.0 & 22.1 & 25.9 & 25.2 & 50.2 & -    & 52.0 & 52.4 & \textbf{80.9} & 70.2                              \\
        \textbf{glue}   & 38.5 & 35.8 & 39.6 & 32.4 & 49.6 & -    & 51.4 & 53.8 & \textbf{68.1} & 60.1                              \\
        holepuncher     & 22.1 & 36.0 & 21.3 & 49.5 & 39.7 & -    & 45.6 & 54.2 & 74.7          & \textbf{85.9}                     \cr\hline
        MEAN            & 24.9 & 27.0 & 27.0 & 32.0 & 40.8 & 47.3 & 43.3 & 47.5 & 63.2 & \textbf{66.2}
        \cr\hline  
    \end{tabular}
    \caption{Quantitative evaluation on the Occlusion-LineMOD dataset. The ADD-0.1d \cite{hinterstoisser2012model} metric is reported and symmetric objects are in bold.}
    \label{tab:OCC_LM_PFM}
\end{table*}

\subsection{Implementation: Different Representation Learning Frameworks}
In this section, we demonstrate the implementation details of different representation learning frameworks in Table $4$. To implement CNN-R$\oplus$D, we lift each pixel in the depth image to its corresponding 3D point to get the XYZ map as well as the normal map. We then concatenate them with the RGB map and feed it into a ResNet34-PSPNet encoding-decoding network for feature extraction of each point (pixel). For PCN-R$\oplus$D, we append RGB values of each point to its 3D coordinate as well as its normal vector and then utilize the RandLA-Net for representation learning. The CNN-R+CNN-D utilizes two ResNet34-PSPNet networks for feature extraction from the RGB image and the XYZ and normal maps respectively. Bidirectional fusion modules (Section 3.2) are added to each encoding and decoding layer. For PCN-R+PCN-D, we leverage one RandLA-Net to extract features from the RGB value of each point and another one for representation learning of the 3D coordinate and the normal vector of each point. In the network flow, bidirectional fusion modules are added to each layer for information communication as well. To implement CNN-R+3DC-D, we replace the RandLA-Net with a 3D convolution neural network. In the encoding stages, the voxel size decreases from $32^3$ to $4^3$ ($32^3 \rightarrow 16^3 \rightarrow 8^3 \rightarrow 4^3$) and the feature dimensions increases from $32$ to $256$ ($32\rightarrow 64 \rightarrow 128 \rightarrow 256$). In the decoding stage, the voxel size increases and feature dimensions decrease inversely. Finally, the geometry feature of each point is obtained within a trilinear interpolation manner, as in PVCNN\cite{liu2019pvcnn}, which is then concatenated with the appearance feature from CNN.

\subsection{More Results}

\subsubsection{Quantitative result on the LineMOD dataset.}
More results of 6D pose estimation on the LineMOD dataset are shown in Table \ref{tab:LM_PFM}.

\subsubsection{Quantitative result on the Occlusion-LineMOD dataset.}
We report more results on the Occlusion-LineMOD dataset in Table \ref{tab:OCC_LM_PFM}. We follow the state-of-the-art to train our model on the LineMOD dataset and only use this dataset for testing.

\begin{figure*}
    \centering
    \includegraphics[scale=0.55]{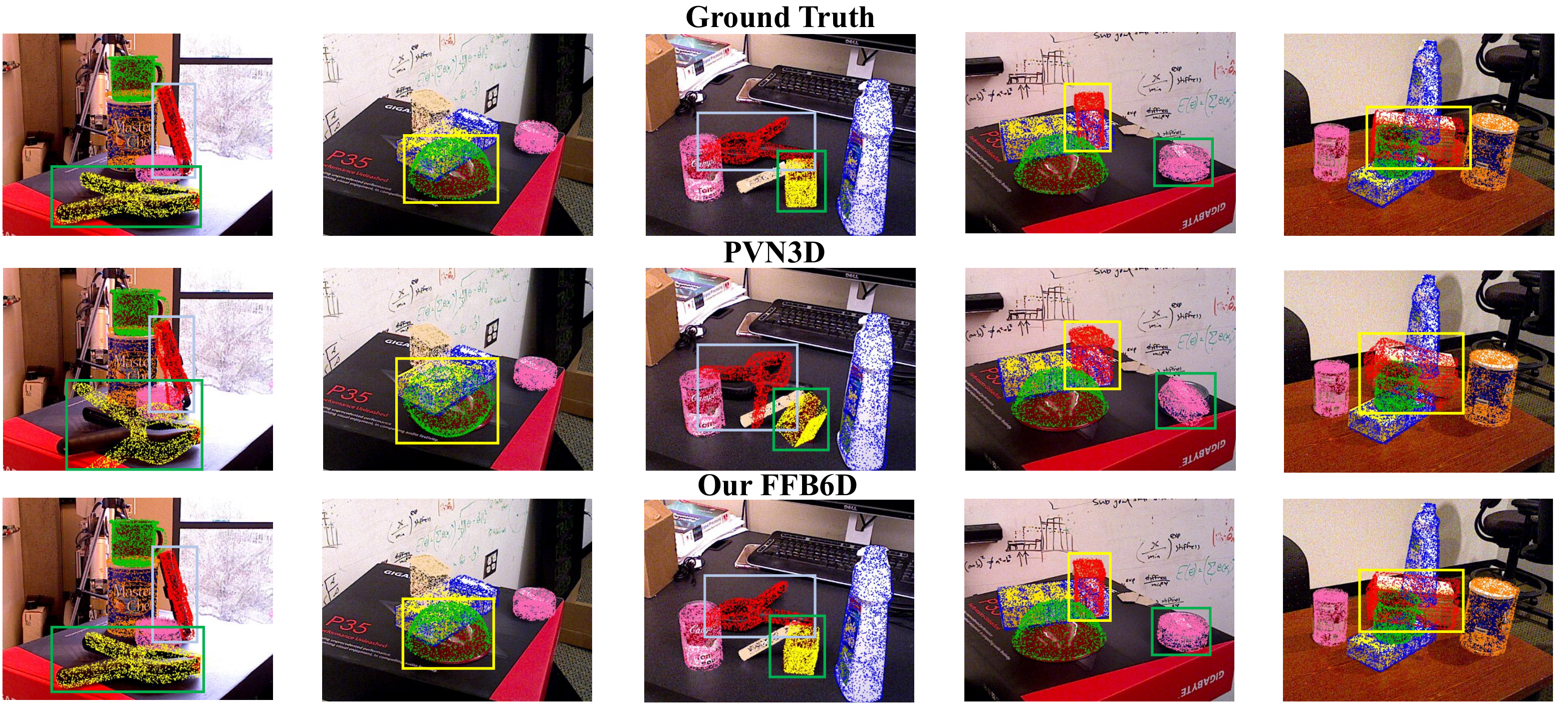}
    \caption{
        Qualitative results of 6D pose on the YCB-Video dataset. Objects in bounding boxes show the pose that we outperform the state-of-the-art significantly. Object vertexes in the object coordinate system are transformed by the ground truth or predicted pose to the camera coordinate system and then projected to the image by the camera intrinsic matrix. Compared to PVN3D \cite{he2020pvn3d} with the DenseFusion \cite{wang2019densefusion} architecture, our FFB6D is more robust towards occlusion and objects with similar appearance or reflective surfaces, which are quite challenging for either isolated CNN or point cloud network feature extraction.
    }
    \label{fig:cmp_pose}
\end{figure*}

\subsubsection{Visualization on predicted pose on the YCB-Video Dataset.}
We provide some qualitative results on the YCB-Video dataset in Figure \ref{fig:cmp_pose}.

{\small
\bibliographystyle{ieee_fullname}
\bibliography{ref}
}

\end{document}